\documentclass[11pt,a4paper]{article} % For LaTeX2e

\usepackage[left=2cm,right=2cm,top=2.5cm,bottom=2.5cm]{geometry}

\usepackage[utf8]{inputenc}
\usepackage[T1]{fontenc}
\usepackage{hyperref}
\usepackage{url}
\usepackage{graphicx}
\usepackage[ruled,linesnumbered]{algorithm2e}
\usepackage{amsmath,amssymb,amsfonts,amsthm}
\usepackage{subcaption}
\usepackage{booktabs}
\usepackage{multirow}
\usepackage{makecell}

\usepackage[round,sort&compress]{natbib}

\usepackage{authblk}
\usepackage{etoolbox}

\newcommand{\keywords}[1]{\par\noindent\textbf{Keywords:} #1}

\makeatletter
\renewcommand\AB@affilsepx{\\[1ex]}
\makeatother

\newtheorem{theorem}{Theorem}

\theoremstyle{definition}
\newtheorem{definition}[theorem]{Definition}

\title{Principal Orthogonal Latent Components Analysis (POLCA Net)}

\author[1]{Jose Antonio Martin H.\thanks{ja.martin.h@repsol.com}}
\author[1]{Freddy Perozo\thanks{f.perozo@repsol.com}}
\author[1]{Manuel Lopez\thanks{manuel.lmartin@repsol.com}}
\affil[1]{Repsol Technology Lab}

\begin{document}

\maketitle

\begin{abstract}
	Representation learning is a pivotal area in the field of machine learning, focusing on the development of methods to automatically discover the representations or features needed for a given task from raw data. Unlike traditional feature engineering, which requires manual crafting of features, representation learning aims to learn features that are more useful and relevant for tasks such as classification, prediction, and clustering.	We introduce Principal Orthogonal Latent Components Analysis Network (POLCA Net), an approach to mimic and extend PCA and LDA capabilities to non-linear domains. POLCA Net combines an autoencoder framework with a set of specialized loss functions to achieve effective dimensionality reduction, orthogonality, variance-based feature sorting, high-fidelity reconstructions, and additionally, when used with classification labels, a latent representation well suited for linear classifiers and low dimensional visualization of class distribution as well.
	
\end{abstract}


\section{Introduction}

Representation learning is a pivotal area in the field of machine learning, focusing on the development of methods to automatically discover the representations or features needed for a given task from raw data. Unlike traditional feature engineering, which requires manual crafting of features, representation learning aims to learn features that are more useful and relevant for tasks such as classification, prediction, and clustering. This approach is integral in the performance of deep learning models, where layers of representation are learned hierarchically to capture increasingly abstract features of the data \citep{bengio2013representation}.

The importance of representation learning lies in its ability to make complex data more accessible for machine learning algorithms. By learning meaningful representations, models can improve generalization to unseen data and reduce the reliance on domain-specific knowledge, thus enabling the application of machine learning in more diverse and complex domains \citep{lecun2015deep}. Techniques such as autoencoders, word embeddings, and convolutional neural networks are prime examples of how representation learning has revolutionized tasks in natural language processing, computer vision, and beyond \citep{goodfellow2016deep}.

As the field progresses, advancements in representation learning continue to enhance the capabilities of machine learning models, driving innovation in areas such as transfer learning, where representations learned in one context are adapted for use in another, and in unsupervised learning, where representations are learned without explicit labels \citep{radford2021learning}. These developments underscore the growing significance of representation learning in shaping the future of artificial intelligence.

Disentangled representations aim to encode information such that different dimensions of the representation correspond to distinct and independent factors of variation, making the learned features more interpretable and useful for downstream tasks. Techniques like $\beta$-VAE (beta-Variational Autoencoders,~\citealt{higgins2017beta}) have been proposed to encourage disentanglement. However, the concepts of disentangled or uncorrelated features seems to be vague or not rigorously or conventionally well defined in some contexts. The lack of formal definitions and standardized evaluation metrics for disentanglement poses challenges for advancing the field~\citep{locatello2019fairness,locatello2019challenging}.

From the formal mathematical point of view, there is a hierarchy~(see appendix~\ref{appedix:disentagled_hierachy}) that begins with the weakest form, linear independence. Linear independence refers to a set of features or functions where no feature can be expressed as a linear combination of the others~\citep{halmos1958finite}. Orthogonality, where features or functions are not only linearly independent but also perpendicular, meaning their inner product is zero~\citep{strang1993introduction}. Finally, the most stringent level is functional independence, where no feature or function can be expressed as a function of any combination of the others, capturing a stronger notion of independence that goes beyond linear relationships~\citep{kolmogorov1950foundations}. This concept is critical in contexts where complete disentanglement of factors of variation is required, such as in the study of complex systems and high-dimensional data. Mutual Information is a measure of Statistical Independence and is usually used as a proxy to approximate it, but is weaker than Functional Independence~\citep{shannon1948mathematical,cover1991elements,kraskov2004estimating,peng2005feature,meyer2008information}. 

In addition to disentanglement, dimensionality reduction is a key focus within the realm of autoencoders. Autoencoders inherently perform dimensionality reduction by encoding input data into a lower-dimensional latent space before reconstructing it. Variational autoencoders (VAEs) and their extensions are particularly noted for balancing dimensionality reduction with the preservation of important data characteristics, often through the incorporation of probabilistic modeling techniques \citep{kingma2013auto}. Information compression is a well-defined concept, particularly in the context of lossless compression, where it refers to the exact recovery of original data from the compressed form. The theoretical limits and challenges, including the undecidability of optimal compression in in the general case, are well-established in this domain~\citep{chaitin1974}. In lossy compression, while the concept remains well-defined, it becomes context-dependent, balancing between compression rate and acceptable loss of quality~\citep{cover1999}.

Among these objectives, Principal Component Analysis (PCA)~\citep{FRS1901LIIIOL,Hotelling1933AnalysisOA} has long been a cornerstone method outside the realm of neural networks, widely adopted for its simplicity, interpretability, and effectiveness in dimensionality reduction. PCA's strengths lie in its ability to reconstruct data from reduced dimensions and its extraction of orthogonal features. In addition to PCA, Linear Discriminant Analysis (LDA) is another powerful technique for dimensionality reduction, particularly in supervised learning contexts. While PCA focuses on maximizing variance to project data into a lower-dimensional space, LDA aims to maximize the separation between different classes by finding a linear combination of features that best separates the classes~\citep{fisher1936use}. This makes LDA particularly effective in classification tasks where the objective is to preserve class separability in the reduced-dimensional space. 

LDA, like PCA, assumes linearity in the data, but it goes a step further by incorporating label information, which PCA does not use. However, LDA's reliance on assumptions such as equal covariance matrices among classes can be a limitation in cases where this assumption does not hold. Despite these limitations, LDA remains a preferred choice for dimensionality reduction when the goal is to enhance class separability rather than just reduce the data's dimensionality.

Here, we introduce Principal Orthogonal Latent Components Analysis Network (POLCA Net), a deep learning architecture designed to capture the benefits of PCA and (optionally) LDA while leveraging non-linear mappings to better handle complex data. POLCA Net extends autoencoders reconstruction loss by incorporating specific constraints as a set of carefully designed complimentary loss functions. POLCA Net enables accurate data reconstruction from reduced dimensions, which is crucial for effective data compression and noise reduction applications. The non-linear nature of POLCA Net allows it to capture more complex relationships in the data, potentially leading to more accurate reconstructions than linear PCA and LDA, especially for datasets with inherent non-linear structures. 

The key features of POLCA Net latent space representation can be summarized as follows:

\begin{enumerate}
	\item Orthogonal latent features: enforces orthogonality in the latent space through a specialized loss term that minimizes the average squared cosine-similarity of the latent components.
	\item Data Compression and dimensionality reduction: encourages the information compression in earlier latent dimensions via a center of mass loss.
	\item Optional Learning with labels: can be trained with class labels to obtain similar functionality to LDA.
	\item Optional Linear decoder: employs a pure linear decoder to maintain theoretical guarantees associated with linear methods, preserves additivity and homogeneity in the latent space, allowing for meaningful algebraic operations on the learned representations.
\end{enumerate}

The experimental results indicate that POLCA Net not only captures the key advantages of PCA and LDA but also provides a versatile alternative for handling complex, high-dimensional data. The agnosticism with respect to the encoder and decoders used and the non-linear capabilities of POLCA Net combined with its ability to maintain PCA-like properties, makes it a powerful tool for modern data analysis and machine learning tasks, bridging the gap between traditional linear techniques and the broad flexibility of deep learning approaches.

\subsection{Related works: PCA, autoencoders, and other dimensionality reduction techniques}

A thorough review of the autoencoder architecture and its variants is presented in by \citep{LI2023110176}. However, the paper notes that there is no clear attempt to reproduce the capabilities of PCA (Principal Component Analysis) in an autoencoder, such as orthogonality and variance sorting in the latent space. The only exception to this is the Kernel-PCA approach, as discussed in \citep{8122623,8517169,Majumdar2021}. Kernel-PCA achieves non-linearity by applying a kernel transformation instead of an activation function.

\paragraph{Orthogonality and Independence} \cite{ren2021learnable} investigate an alternative method for achieving PCA using an autoencoder. Specifically, they aim to achieve results similar to Kernel-PCA by designing a series of linear and nonlinear layers to map the input data into a high-dimensional latent space. To ensure orthogonality in the latent space, the authors use a Cayley transform. However, unlike traditional PCA, the authors do not obtain a set of latent features sorted by data importance. The authors apply the method to the TE process: an industrial benchmark widely applied for the simulation and verification of process monitoring methods.

In the same vein, \cite{COA} proposes a clustering algorithm that employs an autoencoder with an orthogonal constraint in the latent space. However, this approach does not account for the feature ordering based on their significance. The authors ensured the orthogonality constraint by introducing an additional loss term that minimizes the discrepancy between the identity matrix and the multiplication of latent vectors (in a batch) with their transpose. The authors conducted an examination of the challenges associated with imposing this constraint using their technique and comparable methods. The researchers utilize the proposed model on three distinct datasets comprising of images of faces and handwritten digits: MNIST, USPS, and YTF datasets.

\cite{plaut2018principal} implemented PCA using a linear autoencoder and demonstrated that it is possible to extract the PCA loading vectors from the autoencoder weights. However, this method is limited to linear autoencoders. Previously, \cite{Qiu2012} presented an earlier work that applies theoretical approaches to implement PCA using neural networks. The authors introduced Hebbian learning rules and a complex domain extension. They proposed an autoassociative MultiLayer Perceptron (MLP) for nonlinear PCA, which is presented in the study. A similar approach is taken by \cite{978}. 

An insightful survey work presented by \cite{2022factor}, examines the interplay between PCA, Factor Analysis, and Variational Autoencoder (VAE). The relationship between PCA and $\beta$-VAE is noteworthy~\citep{2019variational}, as the regularization with the diagonal normal distribution in VAEs and the feature disentanglement provided by the $\beta$-VAE variant lead to orthogonality. Our method, on the other hand, achieves similar objectives through a distinct and efficient approach.

Also, \cite{huang2018decorrelated}  introduce a batch normalization layer that not only normalizes the inputs to the subsequent layer but also removes correlation between them. To accomplish this, they employ a Zero-Phase Component Analysis (ZCA) whitening matrix and offer an algorithm for its differentiation during backpropagation. They do not present a variance ordering of the outputs nor dimensionality reduction.

\paragraph{Orthogonality and Dimensionality Reduction} \cite{Migenda2021} presents an online neural network-based algorithm for principal component analysis (PCA). The algorithm is complex and includes an eigenvalue approximation inside one of the defined layers. Its purpose is to be practical when an online adjustment is required. The approach is applied on a wide range of datasets with varying characteristics.

In a recent study \cite{Pham2022} and \cite{ladjal2019pcalike} propose a PCA-Autoencoder method for producing independent latent components with dimensions sorted by importance of the data. The method is applied to two sets of images: the CelebA dataset and a custom set of ellipse images. To ensure independence of the latent space components, the authors define and minimize a loss term associated with the magnitude of the covariance matrix during training, in a way similar to the method proposed here. To sort the latent components by feature importance, they propose a method based on training a series of autoencoders, where each successive autoencoder has a larger latent dimension while keeping the previously learned dimensions fixed. This approach is computationally intensive and does not allow for precise control over the actual importance of each latent component, as it assumes a natural adjustment of feature importance by the training process. By contrast, the approach presented here directly manages the ordering of the latent components according to their variance magnitude, enabling greater control over the sorting of components by importance. Additionally, our method avoids the need for training a series of autoencoders, which results in more meaningful and interpretable representations. 

\paragraph{Anomaly and Out-of-Distribution detection} Our proposed solution provides an alternative to traditional anomaly detection techniques, such as Kernel PCA \citep{yang2022generalized}, which rely on sparse representation reconstruction-based methods and entail significant computational overhead. Furthermore, it serves as an alternative to reconstruction-error methods that typically employ autoencoders (AEs), variational autoencoders (VAEs), and generative adversarial networks (GANs, \citealt{yang2022generalized}, \citealt{salehi2022unified} and \citealt{chalapathy2019deep}). Our approach offers the added benefit of an orthogonal and variance-ranked latent space, which further enhances the interpretability and efficiency of the learned representations.

\paragraph{Reduced Order Models (ROMs)} Efforts to develop efficient data-driven Reduced Order Modeling (ROM) techniques, which can create computationally inexpensive lower-order representations of higher-order dynamical systems, have gained considerable attention in recent years \citep{Vinuesa_2022}. An area of particular interest lies in the utilization of neural network-based dimensionality reduction techniques, such as AEs, VAEs, and GANs as well as variants of PCA, as implementation alternatives for reduced order models~\citep{Pant_2021,fenrg_2023_112820,AVERSANO2019422}. POLCA-Net presents a promising alternative to these methods, providing a highly expressive ROM technique that leverages the useful properties of PCA while incorporating non-linearity. Unlike other methods that often generate non-orthogonal and unranked latent spaces, POLCA-Net provides a more computationally efficient solution without sacrificing accuracy, making it well-suited for a variety of applications.

\section{POLCA Net}

\begin{figure}[tb]
	\centering
	\includegraphics[width=0.6\columnwidth]{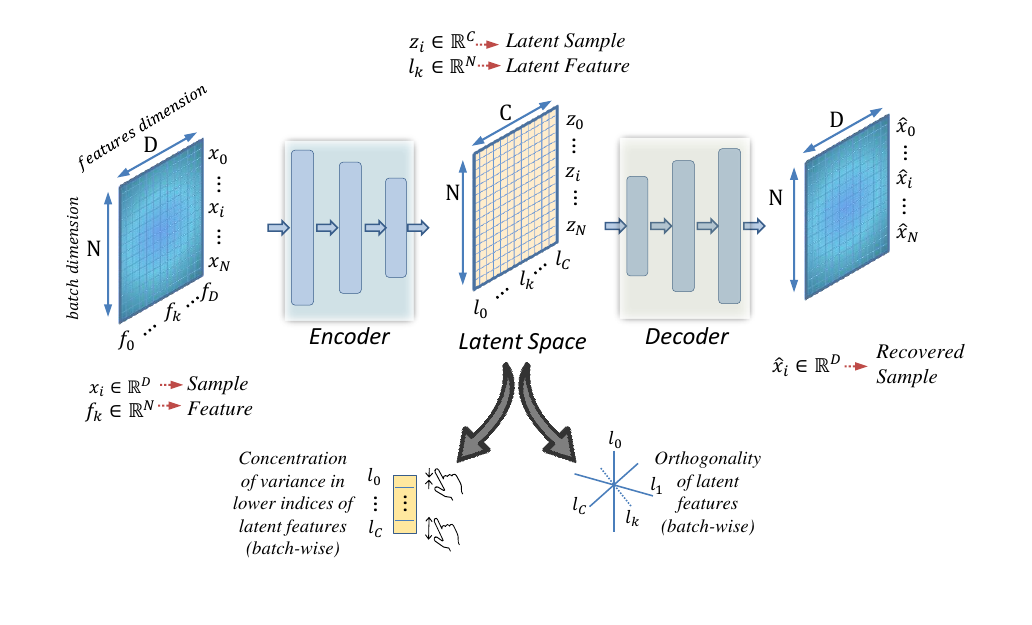}
	\caption{A general autoencoder setting with POLCA Net as the central (bottleneck) component.}	
	\label{fig:model}
\end{figure}

POLCA Net is essentially an autoencoder architecture containing encoder network and a decoder network as shown in Fig.~\ref{fig:model} plus a loss function composed by a set of weighted sub losses,  each defined for a particular purpose or constraint of the latent space desired characteristics. The POLCA's loss acts on the bottleneck part (latent space) generated by the encoder combined with the reconstruction loss which is obtained from the difference between the input to the encoder and the output of the decoder. In terms of dimensionality reduction, POLCA Net reduces the dimensionality of high-dimensional data while preserving its underlying structure and relationships. This compression sorts as well the latent dimensions by their variance, similar to how PCA orders its principal components. This is accomplished through the center of mass loss that encourages the concentration of information in earlier latent dimensions, and optionally allowing for effective truncation of less significant dimensions. Also, a key feature of POLCA Net is its ability to extract orthogonal features from the input data, mirroring one of PCA's most valuable properties. This orthogonality is achieved through a specific term in the loss function that minimizes the cosine similarity between different latent dimensions. Unlike PCA, which achieves a near perfect (up to numeric precision) orthogonality through linear transformations, POLCA Net enforces orthogonality in a non-linear latent space, offering a more flexible and potentially more powerful representation. In addition, POLCA Net can be used in a supervised setting in the same way as LDA, by incorporating an additional classification loss (such as cross entropy) and using the same latent space features as classification variable.

\subsection{Multiobjetive Loss Function:}

POLCA Net uses a composite loss function $\mathcal{L}_\text{polca}$ that guides the learning process:

\begin{align*}
	\mathcal{L}_\text{polca} &= \mathcal{L}_\text{rec} + \mathcal{L}_\text{class} + \alpha \mathcal{L}_\text{ort} + \beta \mathcal{L}_\text{com} + \gamma \mathcal{L}_\text{var},
\end{align*}

where $\mathcal{L}_\text{rec}$ is the reconstruction loss (mean squared error), $\mathcal{L}_\text{class}$ is an optional classification loss (such as cross entropy) or zero, $\mathcal{L}_\text{ort}$ is the orthogonality enforcing loss, $\mathcal{L}_\text{com}$ is the center of mass loss (dimensionality reduction), $\mathcal{L}_\text{var}$~is a variance regularization loss. The hyperparameters $\alpha$, $\beta$ and $\gamma$ are used to weigh each loss component.

The reconstruction loss, $\mathcal{L}_\text{rec}$, is the mean squared error (MSE) between the input and the reconstruction.

\paragraph{Orthogonality and Independence.} The loss $\mathcal{L}_\text{ort}$, encourages the latent features to minimize the cosine similarity matrix $\mathbf{S}$ over the normalized latent representations. The implementation only requires computing the upper triangular matrix due to symmetry excluding the main diagonal as well:

\begin{align*}
\mathcal{L}_\text{ort} &= \frac{2}{n(n-1)} \sum_{1 \leq i < j \leq n} S_{ij}^2
\end{align*}
where $S_{ij}$ is the $(i,j)$-th element of $\mathbf{S} = \mathbf{Z}^T\mathbf{Z}$, and $\mathbf{Z} = [\mathbf{\tilde{z_1}}, \ldots, \mathbf{\tilde{z_N}}]^T$, $\mathbf {\tilde{z_i}} = \frac{\mathbf{z}_i}{\|\mathbf{z}_i\|_2}$, $N$ is the batch size, $n$ the latent space dimension and $\mathbf{z}_{i}$ are the latent features.

\paragraph{Dimensionality Reduction and Variance Regularization.} The center of mass loss, $\mathcal{L}_\text{com}$, is designed to concentrate information in the earlier latent dimensions. It is computed as the weighted average of $L_1$-normalized variances and slightly exponentiated location components $i^{1+\epsilon}$ ($\epsilon=0.25$). The effect of this loss enables progressive reconstruction, where a rough approximation of the input can be obtained from just the first few latent dimensions, with finer details added as more dimensions are included. The variance regularization loss $\mathcal{L}_\text{var}$, control the total per batch variance to prevent a possible gaming against the center of mass loss:
\begin{align*}
        \mathcal{L}_\text{com} &= \frac{\sum_{i=0}^{n-1} i^{1+\epsilon} \cdot \mathbb{E}[(z_i - \mathbb{E}[z_i])^2]}{n \cdot \mathcal{L}_\text{var}}, \\
        \mathcal{L}_\text{var} &= \sum_{i=0}^{n-1} \mathbb{E}[(z_i - \mathbb{E}[z_i])^2]
\end{align*}

\section{Experimental Analysis}

For the experimental work, we used 16 different and diverse datasets containing both gray-scale and color images which are publicly available and widely known. Specifically, we employed the MNIST dataset~\citep{lecun1998gradient} and the FashionMNIST (fmnist, \citealt{fmnist}) dataset available in Pytorch. The 12 2D-MedMNIST~\citep{medmnistv1,medmnistv2} datasets, a large-scale MNIST-like collection of standardized biomedical images, and finally we included two more synthetic datasets generated from simple and high frequency sinusoidal images.

The experiments where performed as a comparison of PCA vs. POLCA Net, by training and testing both on all the refereed datasets. We evaluated the reconstruction capability as well as the classification informativeness of the generated reduced latent space.  For evaluation of reconstruction quality across all datasets we used three standardized metrics: Normalized Mean Square Error (NRMSE), Peak Signal to Noise Ration (PSNR) and the Structural Similarity Index Measure (SSIM) of decoder outputs (reconstruction) vs. original images. For evaluating the classification power of the learnt latent features representation, we trained four different linear classifiers per dataset: The Perceptron, Ridge Classifier, Logistic Regression and Linear SVM (linear kernel), and evaluated the Accuracy and F1-Score metrics, for each classifier and dataset. We used the already defined dataset's train and test split data and collected all the reconstruction and classification metrics for each split as well.

For evaluating the POLCA Net multiobjective loss function, we designed a procedure to collect and analyse the gradients of all the POLCA Net losses during training, to evaluate and validate the interactions between each pair of losses. We use simple definition of loss similarity $(s)$ to evaluate possible loss conflicts or collaborations, the index $s$ is based on the cosine similarity of gradient losses as shown next:
\begin{align}
\label{eq:loss_similairty}
s =\frac{\nabla L_i \cdot \nabla L_j}{|\nabla L_i| |\nabla L_j|}; \; s < {-0.01}~(\text{conflict})
\end{align}

\begin{figure}[tbp]
	\centering
	\begin{subfigure}{.48\textwidth}
		\centering
    	\includegraphics[width=1.0\linewidth]{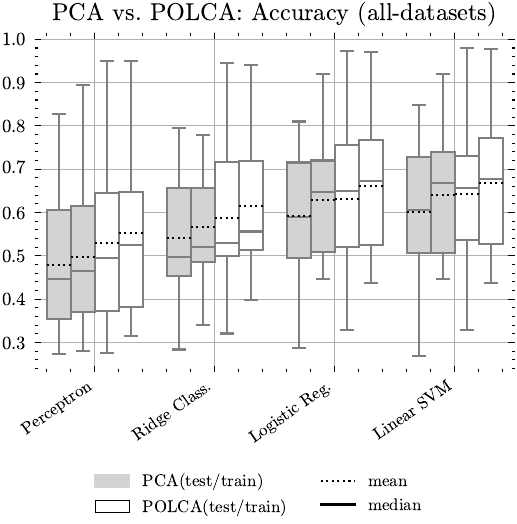}
    	\caption{Classification accuracy distribution.}
    	\label{fig:classifier_metric_comparison_plot}
	\end{subfigure}
	\hfill
	\begin{subfigure}{.48\textwidth}
		\centering
		\includegraphics[width=1.0\linewidth,scale=0.8]{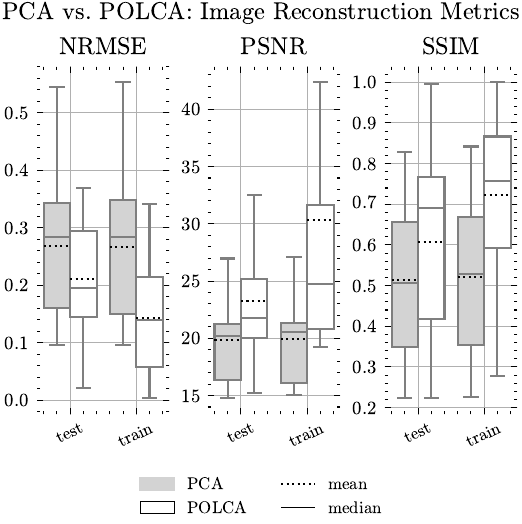}
    	\caption{Image metrics distribution.}
    	\label{fig:image_metrics_comparison_plot}
	\end{subfigure}%
	\caption{Comparison of PCA and POLCA performance across all datasets. (a) Distribution of classification accuracy for PCA and POLCA for each classifier: Perceptron, Ridge Classifier, Logistic Regression, and Linear SVM. (b) Distribution of image reconstruction metrics (NRMSE, PSNR, SSIM) for PCA and POLCA, showing their relative performance in compression and reconstruction.}
	\label{fig:global_results}
\end{figure}

\begingroup
\setlength{\tabcolsep}{2pt} % Default value: 6pt
\begin{table}[htbp]
\caption{Classification Metrics averaged across all Datasets and Image Reconstruction Metris:
    Normalized Root-Mean-Square (NRMSE), Peak Signal to Noise Ratio (PSNR), Structural Similarity Index Metric (SSIM),
    Accuracy and the F1-score-Score.}
\label{tab:joit_classifier_metrics_comparison}
 \centering 
 \small

\begin{tabular}{lcrrlrlrlrlrl}
\toprule
 \multicolumn{13}{c}{Hyperparameters: $\alpha=1\mathrm{e}{-2}$, $\beta=1\mathrm{e}{-2}$, $\gamma=1\mathrm{e}{-7}$ } \\
\midrule
Dataset & \makecell{Latent\\size} &  & \multicolumn{2}{c}{NRMSE} & \multicolumn{2}{c}{PSNR} & \multicolumn{2}{c}{SSIM} & \multicolumn{2}{c}{Accurary} & \multicolumn{2}{c}{F1-score} \\
 &  &  & PCA & POLCA & PCA & POLCA & PCA & POLCA & PCA & POLCA & PCA & POLCA \\
\midrule
\multirow[c]{2}{*}{sinusoidal} & \multirow[c]{2}{*}{8} & test & 0.19 & 0.01 & 20.60 & 46.13 & 0.84 & 1.00 &  &  &  &  \\
 &  & train & 0.19 & 0.00 & 20.51 & 54.50 & 0.84 & 1.00 &  &  &  &  \\
\cmidrule{4-13}
\multirow[c]{2}{*}{bent} & \multirow[c]{2}{*}{46} & test & 0.33 & 0.03 & 14.71 & 36.14 & 0.61 & 1.00 &  &  &  &  \\
 &  & train & 0.31 & 0.00 & 15.29 & 52.31 & 0.67 & 1.00 &  &  &  &  \\
\cmidrule{4-13}
\multirow[c]{2}{*}{mnist} & \multirow[c]{2}{*}{11} & test & 0.55 & 0.30 & 15.13 & 20.45 & 0.50 & 0.80 & 0.77 & 0.95 & 0.76 & 0.95 \\
 &  & train & 0.55 & 0.30 & 15.08 & 20.62 & 0.50 & 0.81 & 0.76 & 0.94 & 0.75 & 0.94 \\
\cmidrule{4-13}
\multirow[c]{2}{*}{fmnist} & \multirow[c]{2}{*}{8} & test & 0.40 & 0.28 & 16.19 & 19.65 & 0.47 & 0.67 & 0.69 & 0.74 & 0.67 & 0.74 \\
 &  & train & 0.40 & 0.28 & 16.18 & 19.76 & 0.47 & 0.68 & 0.69 & 0.75 & 0.67 & 0.74 \\
\cmidrule{4-13}
\multirow[c]{2}{*}{breast} & \multirow[c]{2}{*}{8} & test & 0.26 & 0.30 & 20.46 & 21.27 & 0.36 & 0.43 & 0.75 & 0.74 & 0.71 & 0.68 \\
 &  & train & 0.26 & 0.01 & 20.60 & 55.23 & 0.38 & 1.00 & 0.74 & 0.73 & 0.72 & 0.68 \\
\cmidrule{4-13}
\multirow[c]{2}{*}{derma} & \multirow[c]{2}{*}{8} & test & 0.10 & 0.09 & 24.84 & 25.45 & 0.69 & 0.71 & 0.64 & 0.66 & 0.56 & 0.56 \\
 &  & train & 0.10 & 0.08 & 24.86 & 26.29 & 0.69 & 0.72 & 0.64 & 0.66 & 0.55 & 0.56 \\
\cmidrule{4-13}
\multirow[c]{2}{*}{oct} & \multirow[c]{2}{*}{8} & test & 0.36 & 0.23 & 20.20 & 24.42 & 0.49 & 0.68 & 0.29 & 0.33 & 0.19 & 0.24 \\
 &  & train & 0.38 & 0.23 & 20.72 & 25.07 & 0.54 & 0.75 & 0.48 & 0.50 & 0.41 & 0.45 \\
\cmidrule{4-13}
\multirow[c]{2}{*}{organa} & \multirow[c]{2}{*}{8} & test & 0.34 & 0.34 & 15.92 & 15.80 & 0.22 & 0.24 & 0.51 & 0.55 & 0.49 & 0.53 \\
 &  & train & 0.34 & 0.26 & 15.63 & 17.78 & 0.23 & 0.37 & 0.57 & 0.62 & 0.55 & 0.59 \\
\cmidrule{4-13}
\multirow[c]{2}{*}{organc} & \multirow[c]{2}{*}{8} & test & 0.32 & 0.32 & 16.40 & 16.33 & 0.25 & 0.28 & 0.54 & 0.65 & 0.51 & 0.64 \\
 &  & train & 0.32 & 0.25 & 15.91 & 17.75 & 0.24 & 0.35 & 0.59 & 0.70 & 0.56 & 0.69 \\
\cmidrule{4-13}
\multirow[c]{2}{*}{organs} & \multirow[c]{2}{*}{8} & test & 0.31 & 0.32 & 16.67 & 16.38 & 0.25 & 0.26 & 0.36 & 0.37 & 0.31 & 0.32 \\
 &  & train & 0.31 & 0.26 & 16.24 & 17.60 & 0.24 & 0.32 & 0.39 & 0.40 & 0.34 & 0.36 \\
\cmidrule{4-13}
\multirow[c]{2}{*}{path} & \multirow[c]{2}{*}{8} & test & 0.16 & 0.16 & 20.17 & 20.11 & 0.30 & 0.30 & 0.47 & 0.45 & 0.42 & 0.42 \\
 &  & train & 0.15 & 0.15 & 20.85 & 20.78 & 0.27 & 0.27 & 0.43 & 0.49 & 0.38 & 0.44 \\
\cmidrule{4-13}
\multirow[c]{2}{*}{pneumonia} & \multirow[c]{2}{*}{8} & test & 0.13 & 0.14 & 22.58 & 22.60 & 0.61 & 0.65 & 0.80 & 0.79 & 0.79 & 0.77 \\
 &  & train & 0.13 & 0.10 & 22.74 & 24.60 & 0.61 & 0.69 & 0.91 & 0.92 & 0.91 & 0.92 \\
\cmidrule{4-13}
\multirow[c]{2}{*}{retina} & \multirow[c]{2}{*}{8} & test & 0.15 & 0.15 & 26.97 & 27.34 & 0.82 & 0.84 & 0.48 & 0.52 & 0.40 & 0.49 \\
 &  & train & 0.14 & 0.08 & 27.10 & 31.65 & 0.83 & 0.91 & 0.49 & 0.51 & 0.42 & 0.48 \\
\cmidrule{4-13}
\multirow[c]{2}{*}{blood} & \multirow[c]{2}{*}{8} & test & 0.14 & 0.13 & 19.95 & 20.32 & 0.51 & 0.54 & 0.57 & 0.61 & 0.56 & 0.59 \\
 &  & train & 0.14 & 0.12 & 19.97 & 20.99 & 0.52 & 0.56 & 0.57 & 0.60 & 0.56 & 0.58 \\
\cmidrule{4-13}
\multirow[c]{2}{*}{chest} & \multirow[c]{2}{*}{8} & test & 0.18 & 0.14 & 20.81 & 22.70 & 0.67 & 0.75 & 0.49 & 0.48 & 0.02 & 0.02 \\
 &  & train & 0.18 & 0.14 & 20.76 & 22.73 & 0.67 & 0.75 & 0.49 & 0.49 & 0.02 & 0.02 \\
\cmidrule{4-13}
\multirow[c]{2}{*}{tissue} & \multirow[c]{2}{*}{8} & test & 0.39 & 0.37 & 26.91 & 27.47 & 0.65 & 0.68 & 0.41 & 0.42 & 0.33 & 0.32 \\
 &  & train & 0.39 & 0.37 & 26.93 & 27.48 & 0.65 & 0.68 & 0.41 & 0.42 & 0.33 & 0.32 \\
\cmidrule{4-13}
\bottomrule
\end{tabular}
\end{table}
\endgroup

\begin{figure}[tbp]
	\centering
    \includegraphics[width=1.0\linewidth]{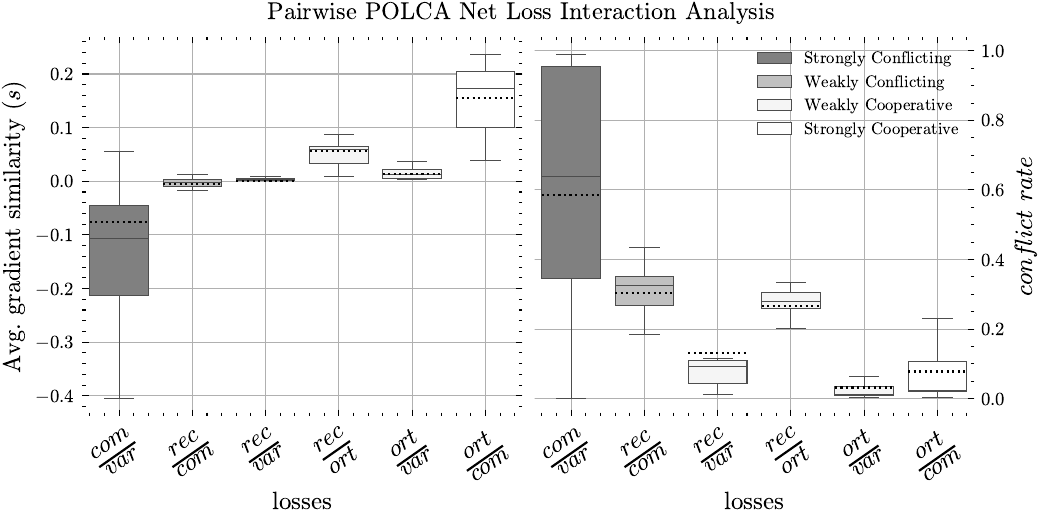}
	\caption{Pairwise analysis of POLCA multiobjective loss statistics: reconstruction loss ($L_{rec}$), orthogonality loss ($L_{ort}$), dimensionality reduction center of mass loss ($L_{com}$) and variance regularization loss ($L_{var}$), collected during training phase on all the experiments realized. The gradient similarity $(s)$ and conflicts are defined in Equation~\ref{eq:loss_similairty}}
	\label{fig:loss_conflicts}
\end{figure}

In classification tasks, Table~\ref{tab:joit_classifier_metrics_comparison} and Fig.~\ref{fig:classifier_metric_comparison_plot}, POLCA Net consistently outperforms PCA across all tested linear classifiers. 

In terms of image reconstruction, Table~\ref{tab:joit_classifier_metrics_comparison} and Fig.~\ref{fig:image_metrics_comparison_plot}, POLCA Net demonstrates superior performance across all evaluated metrics. The Normalized Root Mean Square Error (NRMSE) is significantly lower for POLCA Net, indicating better overall reconstruction accuracy. The Structural Similarity Index (SSIM) shows advantage for POLCA Net, suggesting better preservation of structural information. Most notably, the Peak Signal-to-Noise Ratio (PSNR) is substantially higher for POLCA Net, indicating superior reconstruction quality and less noise in the reconstructed images. These results highlight POLCA Net's balanced performance in both classification and reconstruction tasks.

\section{Conclusion}

This study introduced POLCA Net, an autoencoder-based approach for dimensionality reduction and feature extraction. The key findings are:

\begin{enumerate}
    \item POLCA Net combines multiple loss functions to achieve orthogonality, variance-based feature sorting, and dimensionality reduction in the latent space.
    
    \item Experiments were conducted on 16 diverse datasets, including MNIST, FashionMNIST, MedMNIST, and synthetic datasets.
    
    \item Performance was evaluated using reconstruction metrics (NRMSE, PSNR, SSIM) and classification metrics (Accuracy, F1-Score) for four linear classifiers.
    
    \item Results showed that POLCA Net consistently outperformed PCA in classification tasks across all tested linear classifiers.
    
    \item In image reconstruction, POLCA Net demonstrated superior performance across all evaluated metrics, with lower NRMSE, higher SSIM, and higher PSNR compared to PCA.
    
    \item Analysis of the multi-objective loss function revealed interactions between different loss components during training.
    
    \item The study provided a mathematical proof of functional independence for the loss components used in POLCA Net.
    
    \item The experimental setup and implementation details were provided to ensure reproducibility of the results.
\end{enumerate}

These findings suggest that POLCA Net offers an alternative approach to dimensionality reduction and feature extraction, combining aspects of traditional techniques like PCA with the flexibility of neural network-based methods. Further research may be needed to explore its effectiveness in various domains and applications.

\section{Replicability}

General experimental infrastructure and fixed  parameters are shown in Table~\ref{tab:exp_params}.
\begin{table}[h]
\centering
\small
\begin{tabular}{ll}
\toprule
CPU: AMD EPYC 7V13 64-Core (24/24) & Linux: 4.18.0-553.16.1.el8\_10.x86\_64 \\
Device: NVIDIA A100 80GB PCIe &  CUDA Version: 12.4 \\
Python Version: 3.12.5 & PyTorch Version: 2.4.0 \\
Gradient Updates: $20\mathrm{k}$ & Batch Size: 64 \\
Random seed$=5$ for all Python, Numpy and PyTorch\\
\bottomrule
\end{tabular}
\caption{Experimental Parameters}
\label{tab:exp_params}
\end{table}

Full source code of the POLCA Net implementation as well as all the required code  (including scripts and jupyter notebooks) to reproduce the experiments and perform new ones has been provided as a pip installable single python package in the additional material.

\bibliography{references}

\bibliographystyle{plainnat}

\appendix

\section{Hierarchy of Linear Independence, Orthogonality, and Functional Independence} 
\label{appedix:disentagled_hierachy}

\begin{definition}[Linear Independence of Functions, \cite{axler2015linear}]
	A set of functions $\{f_1, f_2, \dots, f_n\}$ defined on a domain $\mathcal{D}$ is linearly independent if:
	$$
	\sum_{i=1}^n \alpha_i f_i(x) = 0 \quad \forall x \in \mathcal{D} \implies \alpha_i = 0 \quad \forall i
	$$
	
\end{definition}

\begin{definition}[Orthogonality of Functions, \cite{debnath2005introduction}]
	A set of functions $\{f_1, f_2, \dots, f_n\}$ is orthogonal with respect to an inner product $\langle \cdot, \cdot \rangle$ if:
	$$
	\langle f_i, f_j \rangle = 0 \quad \forall i \neq j
	$$
	
\end{definition}

\begin{definition}[Functional Independence, \cite{hirsch1976differential, lee2012introduction}]
	A set of functions $\{f_1, f_2, \dots, f_n\}$ is functionally independent if there exists no non-trivial function $\Phi$ such that:
	$$
	\Phi(f_1(x), f_2(x), \dots, f_n(x)) = 0 \quad \forall x \in \mathcal{D}
	$$
	
\end{definition}

Example: Consider the functions $\sin(x)$, $\cos(x)$, and $g(x) = \sin^2(x) + \cos^2(x)$ on $[0, 2\pi]$:
\begin{itemize}
	\item They are linearly independent.
	\item $\sin(x)$ and $\cos(x)$ are orthogonal: $\int_0^{2\pi} \sin(x)\cos(x) dx = 0$
	\item They are not functionally independent because $g(x) = 1$ for all $x$
\end{itemize}

This example illustrates that orthogonality does not imply functional independence, and linear independence does not imply orthogonality.

Thus, functionally independent functions might not be orthogonal as they could have non-zero inner products, but still no function of them equals zero. Orthogonal functions are always linearly independent, but the reverse isn't true. Linear independence is the weakest condition and doesn't guarantee orthogonality or functional independence.

In applications requiring truly independent features or representations, verifying functional independence is necessary, as orthogonality or linear independence alone may not be sufficient to ensure complete independence of the functions or features.

This functional independence implies that each of these functionals captures a unique aspect of the function $f(x)$ that cannot be derived from the others, making them valuable and distinct measures in function analysis. In practical applications, such as signal processing or data analysis, this independence suggests that considering all three measures can provide a more comprehensive understanding of the underlying function or data.

\subsection{Example: Functional Independence of Area, Center of Mass, and Curve Length}

As an illustrative example, lets study three independent functionals used in computer vision~\citep[{although author's claim for orthogonal variant moments the right term shall be functional independent}]{MARTINH2010846}
	
Consider a function $f(x)$ defined on the interval $[a,b]$, where $f(x)$ is continuous on $[a,b]$ and differentiable on $(a,b)$. We will analyze three functionals: the area under the curve $A[f]$, the center of mass $C[f]$, and the curve length $L[f]$.

\begin{enumerate}
	\item \textbf{Area under the curve:}
	$$
	A[f] = \int_a^b f(x) \, dx
	$$
	
	\item \textbf{Center of mass:}
	$$
	C[f] = \frac{1}{A[f]} \int_a^b x f(x) \, dx
	$$
	
	\item \textbf{Curve length:}
	$$
	L[f] = \int_a^b \sqrt{1 + (f'(x))^2} \, dx
	$$
\end{enumerate}

\paragraph{Proof of Functional Independence} To prove that $A[f]$, $C[f]$, and $L[f]$ are functionally independent, we need to show that there exists no non-trivial function $\Phi$ such that:

$$
\Phi(A[f], C[f], L[f]) = 0 \quad \forall f
$$

We will demonstrate this by showing that each functional can be altered independently of the others using specific transformations of $f(x)$.

\paragraph{Independence of $A[f]$ and $C[f]$}

Consider the transformation $T_1[f](x) = f(x) + \epsilon(x - \frac{a+b}{2})$, where $\epsilon$ is a small non-zero constant.

For the area functional:

\begin{align*}
	A[T_1[f]] &= \int_a^b \left(f(x) + \epsilon(x - \frac{a+b}{2})\right) \, dx \\
	&= A[f] + \epsilon \int_a^b \left(x - \frac{a+b}{2}\right) \, dx \\
	&= A[f] + \epsilon \left[\frac{x^2}{2} - \frac{a+b}{2}x\right]_a^b \\
	&= A[f]
\end{align*}

Since the integral of a linear term symmetric around $(a+b)/2$ cancels out, $A[T_1[f]] = A[f]$.

For the center of mass functional:

\begin{align*}
	C[T_1[f]] &= \frac{1}{A[T_1[f]]} \int_a^b x\left(f(x) + \epsilon(x - \frac{a+b}{2})\right) \, dx \\
	&= C[f] + \frac{\epsilon}{A[f]} \int_a^b \left(x^2 - x\frac{a+b}{2}\right) \, dx \\
	&= C[f] + \frac{\epsilon}{A[f]} \left[\frac{x^3}{3} - \frac{a+b}{4}x^2\right]_a^b \\
	&\neq C[f] \text{ for } \epsilon \neq 0
\end{align*}

This shows that $C[f]$ can be altered independently of $A[f]$.

\paragraph{Independence of $L[f]$ from $A[f]$ and $C[f]$}

Consider the transformation $T_2[f](x) = f(x) + \epsilon \sin\left(\frac{2\pi n}{b-a}x\right)$, where $n$ is a large integer and $\epsilon$ is small.

For the area functional:

\begin{align*}
	A[T_2[f]] &= \int_a^b \left(f(x) + \epsilon \sin\left(\frac{2\pi n}{b-a}x\right)\right) \, dx \\
	&= A[f] + \epsilon \int_a^b \sin\left(\frac{2\pi n}{b-a}x\right) \, dx \\
	&= A[f]
\end{align*}

Since the integral of a sine function over its period is zero, $A[T_2[f]] = A[f]$.

For the center of mass functional:

\begin{align*}
	C[T_2[f]] &= \frac{1}{A[f]} \int_a^b x\left(f(x) + \epsilon \sin\left(\frac{2\pi n}{b-a}x\right)\right) \, dx \\
	&= C[f] + \frac{\epsilon}{A[f]} \int_a^b x \sin\left(\frac{2\pi n}{b-a}x\right) \, dx \\
	&= C[f] + O\left(\frac{\epsilon}{n}\right) \quad \text{as } n \to \infty
\end{align*}

The integral of $x\sin\left(\frac{2\pi n}{b-a}x\right)$ tends to zero as $n$ increases, so $C[f]$ is nearly unaffected.

For the curve length functional:

\begin{align*}
	L[T_2[f]] &= \int_a^b \sqrt{1 + \left(f'(x) + \epsilon \frac{2\pi n}{b-a} \cos\left(\frac{2\pi n}{b-a}x\right)\right)^2} \, dx \\
	&\approx L[f] + \frac{\epsilon^2 \left(\frac{2\pi n}{b-a}\right)^2}{2} \int_a^b \cos^2\left(\frac{2\pi n}{b-a}x\right) \, dx \\
	&\approx L[f] + \frac{\epsilon^2 \left(\frac{2\pi n}{b-a}\right)^2}{4}(b-a)
\end{align*}

This shows that $L[f]$ can be significantly altered while $A[f]$ and $C[f]$ remain almost unchanged.

We have demonstrated that:
\begin{itemize}
	\item $A[f]$ can be kept constant while $C[f]$ is altered.
	\item $C[f]$ can be kept nearly constant while $L[f]$ is altered.
	\item $L[f]$ can be significantly altered while $A[f]$ and $C[f]$ remain nearly constant.
\end{itemize}

Thus, there exists no non-trivial function $\Phi$ such that $\Phi(A[f], C[f], L[f]) = 0$ for all $f$. This proves that $A[f]$, $C[f]$, and $L[f]$ are functionally independent.
\qed

\paragraph{Limitations and Considerations} The proof assumes that $f(x)$ is continuous on $[a,b]$ and differentiable on $(a,b)$. The transformations used are local; they demonstrate independence in a neighborhood of any given function $f$. While we've shown that each functional can be changed independently, we haven't explicitly considered all possible combinations of simultaneous changes.

\section{Functional Independence of POLCA Losses}

Consider a POLCA Net based autoendoder with input $x$, latent representation $z$, and output $\hat{x}$. We will analyze three loss functions: variance sorting loss $L_{com}$, variance reduction loss $L_{var}$, and reconstruction loss $L_2$.

\paragraph{Definitions} Let $z_i$ be the $i$-th component of the latent vector $z$, and $\sigma_i^2$ be its variance across a batch of inputs.

\begin{enumerate}
	\item \textbf{Variance Sorting Loss}:
	$$
	L_{com} = \frac{1}{d} \sum_{i=1}^d i \cdot \sigma_{(i)}^2
	$$
	where $\sigma_{(i)}^2$ are the sorted variances in descending order, and $d$ is the dimension of $z$.
	
	\item \textbf{Variance Reduction Loss}:
	$$
	L_{var} = \sum_{i=1}^d \sigma_i^2
	$$
	
	\item \textbf{Reconstruction Loss}:
	$$
	L_2 = \|x - \hat{x}\|_2^2
	$$
\end{enumerate}

\paragraph{Proof of Functional Independence}

To prove that $L_{com}$, $L_{var}$, and $L_2$ are functionally independent, we need to show that there exists no non-trivial function $\Phi$ such that:

$$
\Phi(L_{com}, L_{var}, L_2) = 0 \quad \forall \text{ possible autoencoders}
$$

We will demonstrate this by showing that we can change each loss independently of the others. A key idea behind this proof is that in general ML classification and regression algorithms are blind with respect to features ordering and feature scaling and or normalization is usually performed as a previous step or is integrated inside the algorithm.

\paragraph{Independence of $L_{com}$ and $L_{var}$}

Consider the transformation $T_1[z] = Pz$, where $P$ is a permutation matrix.

For $L_{var}$:
\begin{align*}
	L_{var}(T_1[z]) &= \sum_{i=1}^d \sigma_i^2(P z) \\
	&= \sum_{i=1}^d \sigma_i^2(z) \\
	&= L_{var}(z)
\end{align*}

For $L_{com}$:
\begin{align*}
	L_{com}(T_1[z]) &= \frac{1}{d} \sum_{i=1}^d i \cdot \sigma_{(i)}^2(P z) \\
	&\neq L_{com}(z) \text{ for non-trivial permutations}
\end{align*}

This transformation changes $L_{com}$ while keeping $L_{var}$ constant.

\paragraph{Independence of $L_2$ from $L_{com}$ and $L_{var}$}

Consider the transformation $T_2[z] = \alpha z$, where $\alpha \neq 0$ is a scaling factor.

For $L_{com}$ and $L_{var}$:
\begin{align*}
	L_{com}(T_2[z]) &= \alpha^2 L_{com}(z) \\
	L_{var}(T_2[z]) &= \alpha^2 L_{var}(z)
\end{align*}

For $L_2$, assuming the decoder can compensate for the scaling:
\begin{align*}
	L_2(T_2[z]) &= \|x - \hat{x}(T_2[z])\|_2^2 \\
	&= \|x - \hat{x}(z)\|_2^2 \\
	&= L_2(z)
\end{align*}

This transformation changes $L_{com}$ and $L_{var}$ while keeping $L_2$ constant.

\paragraph{Independence of $L_2$ from $L_{com}$ and $L_{var}$}

Consider the transformation $T_3[\hat{x}] = \hat{x} + \epsilon$, where $\epsilon$ is a small non-zero vector.

For $L_{com}$ and $L_{var}$:
\begin{align*}
	L_{com}(T_3[\hat{x}]) &= L_{com}(\hat{x}) \\
	L_{var}(T_3[\hat{x}]) &= L_{var}(\hat{x})
\end{align*}

For $L_2$:
\begin{align*}
	L_2(T_3[\hat{x}]) &= \|x - (\hat{x} + \epsilon)\|_2^2 \\
	&= \|x - \hat{x}\|_2^2 + \|\epsilon\|_2^2 - 2(x - \hat{x})^T\epsilon \\
	&\neq L_2(\hat{x}) \text{ for } \epsilon \neq 0
\end{align*}

This transformation changes $L_2$ while keeping $L_{com}$ and $L_{var}$ constant.

\paragraph{Conclusion}

We have demonstrated that:
\begin{itemize}
	\item $L_{com}$ can be changed independently of $L_{var}$ and $L_2$ (using $T_1$)
	\item $L_{var}$ can be changed independently of $L_{com}$ and $L_2$ (using $T_1$ and $T_2$)
	\item $L_2$ can be changed independently of $L_{com}$ and $L_{var}$ (using $T_3$)
\end{itemize}

Therefore, there exists no non-trivial function $\Phi$ such that $\Phi(L_{com}, L_{var}, L_2) = 0$ for all possible autoencoders. This proves that $L_{com}$, $L_{var}$, and $L_2$ are functionally independent.
\qed

The independence of these losses suggests that they can be combined in a multi-objective optimization framework to achieve a more comprehensive and fine-tuned autoencoder performance. Each loss provides a distinct signal for optimization, potentially leading to more robust and informative latent representations.

\paragraph{Limitations and Considerations}

1. This proof assumes that the autoencoder has sufficient capacity to compensate for transformations in the latent space.
2. The transformations used are local; they demonstrate independence in a neighborhood of any given autoencoder configuration.
3. In practical implementations, the exact independence may be affected by the specific architecture and optimization procedure of the autoencoder.
4. The proof does not consider potential indirect interactions that might arise in the optimization process when these losses are combined.

\end{document}